\documentclass[11pt]{article}

\usepackage[preprint]{acl}

\usepackage{latexsym}

\usepackage{microtype}

\usepackage{graphicx}
\usepackage{booktabs}
\usepackage{colortbl}
\usepackage{amsmath}
\usepackage{amssymb}
\usepackage{tikz}
\usetikzlibrary{positioning,arrows.meta,fit,calc}
\usepackage{algorithm}
\usepackage{algpseudocode}
\usepackage[no-math]{fontspec}
\newcommand{\method}{Multi-View Latent MoE\xspace}
\newcommand{\mr}{Multi-Router\xspace}
\newcommand{\mv}{Multi-View\xspace}

\usepackage{xspace}

\newcommand{\paramDeltaMV}{$0.19\%$}

\newcommand{\paramDeltaMR}{$0.06\%$}

\usepackage{ernie_frontmatter}
\title{\ErniePaperTitle}
\author{Zelin Zhao\thanks{Project Lead}, Junyuan Shang, Tianchi Ma, Shuohuan Wang\thanks{Corresponding Author}, Dianhai Yu\\ERNIE Team, Baidu}

\begin{document}
\BeginErnieEnglish
  \onecolumn
  \thispagestyle{erniefirst}
  \begingroup
  \parindent=0pt
  \noindent\includegraphics[width=\textwidth,height=5pt]{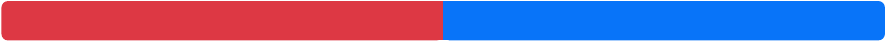}\par
  \vspace{-1pt}
  \noindent\begin{tikzpicture}
    \node[fill=erniebackground,draw=none,rounded corners=6pt,
      inner xsep=0.6cm,inner ysep=0.9cm,
      text width=\dimexpr\textwidth-1.2cm\relax] {%
      \begin{minipage}{\linewidth}
      \ErnieCoverFont
      \centering
      \begin{minipage}{0.90\linewidth}
      {\centering\fontsize{16}{20}\selectfont\bfseries
        \includegraphics[height=20pt]{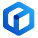}\hspace{7pt}%
        Multi-View Latent MoE:\par
        Letting Every Router Shape Expert Computation\par}
      \vspace{16pt}
      {\fontsize{10.5}{14}\selectfont\centering\bfseries
        Zelin Zhao\textsuperscript{\ensuremath{\dagger}}, Junyuan Shang, Tianchi Ma,\par
        Shuohuan Wang\textsuperscript{*}, Dianhai Yu\par
        \vspace{6pt}ERNIE Team, Baidu\par}
      \vspace{20pt}
      {\fontsize{11}{14}\selectfont\bfseries Abstract\par}
      \vspace{5pt}
      {\fontsize{10.5}{13}\selectfont Sparse mixture-of-experts (MoE) models expand parameter capacity by activating only a small subset of experts for each token. Expert selection directly affects model capabilities under a fixed active-expert budget. However, dense evaluation of all experts reveals weak agreement between single-router score rankings and rankings from expert-utility evaluations, especially in middle and deep layers. Obtaining richer signals for selection and computation entails additional cost: selection based on internal expert activations requires preliminary computation, while processing multiple full representations separately through the same expert repeats its computation. We propose Multi-View Latent MoE to improve expert selection with multiple routers and further reuse their preferences to modulate expert computation under a fixed expert budget. Its Multi-Router submodule aggregates sigmoid scores from independently initialized routers. These routers develop highly differentiated expert preferences during training, allowing experts strongly favored by one router or jointly supported by several routers to be selected. Multi-View further uses each selected expert's relative scores across routers to mix learned latent views into its input, reusing routing preferences for both selection and computation. Each selected expert executes one forward pass, preserving the active-expert count, expert FLOPs, and all-to-all communication volume. On a 24-layer 10B-A1B model trained on 708B tokens, full Multi-View Latent MoE improves the mean score across nine downstream tasks by 1.18 percentage points over the baseline. Its Multi-Router submodule alone yields a 0.6-percentage-point gain and improves the rank correlation between router scores and evaluated expert utility, with relative increases of 8.1\% globally and approximately 58\% in the layer with the lowest single-router alignment. Controlled mixing interventions further show that the model uses input differences that depend on both the expert and the token.
\par}
      \vspace{12pt}
      {\fontsize{9}{11}\selectfont\bfseries\ErniePaperDate\par}
      \end{minipage}
      \end{minipage}};
  \end{tikzpicture}\par
  \endgroup
  \begingroup
    \renewcommand{\thefootnote}{\fnsymbol{footnote}}
    \footnotetext[1]{Corresponding Author.}
    \footnotetext[2]{Project Lead.}
  \endgroup
  \setcounter{footnote}{0}
\clearpage
\twocolumn
\suppressfloats[t]
\begin{figure}[t]
\centering
\includegraphics[width=\columnwidth]{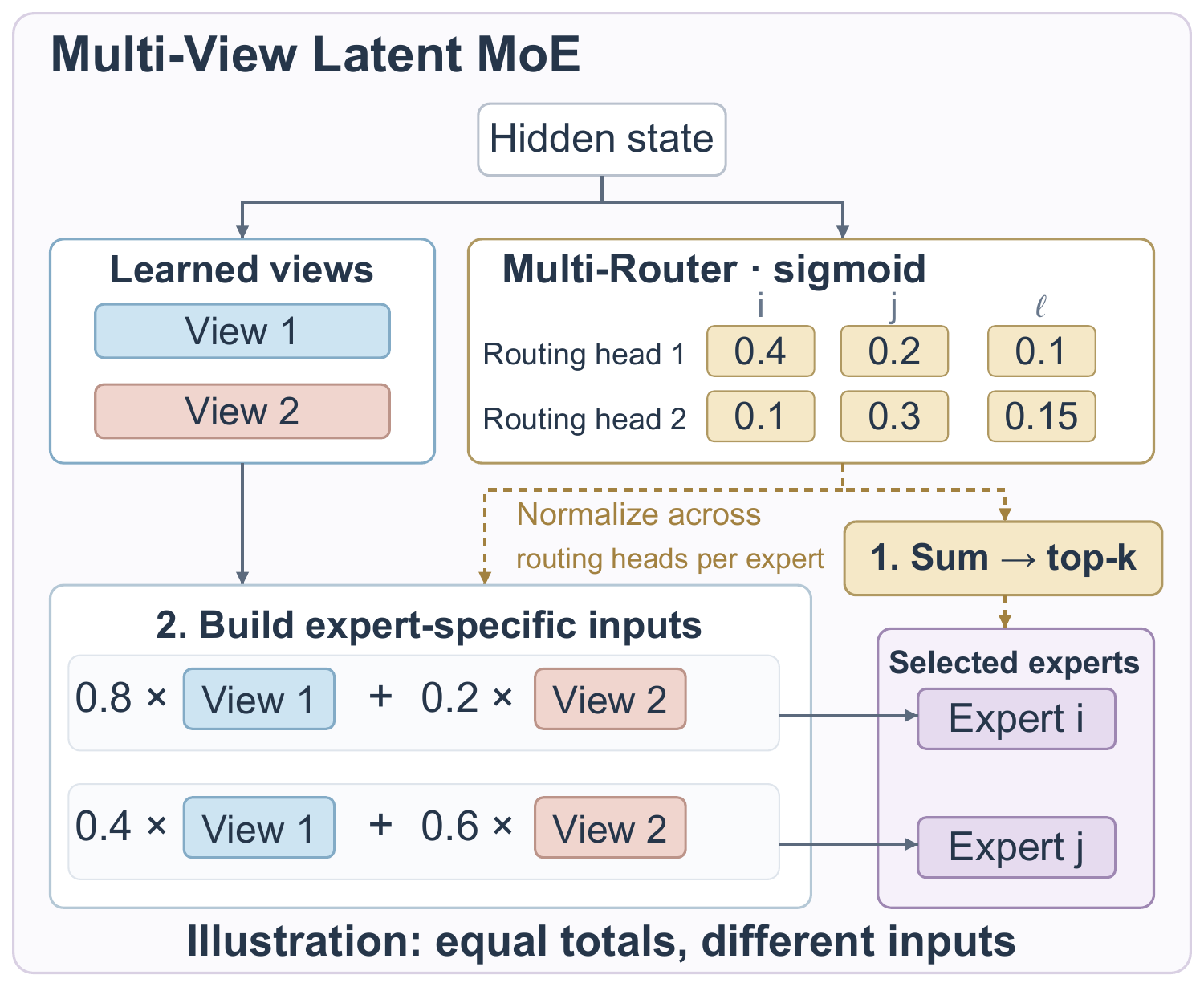}
\caption{Router-conditioned expert inputs in \mv{}. Equal total scores can express different preferences across routing heads and yield different mixed inputs. In this illustrative $k=2$ example, experts $i$ and $j$ have score pairs $(0.4,0.1)$ and $(0.2,0.3)$, both summing to $0.5$. Normalizing across routing heads for each expert gives view proportions $(0.8,0.2)$ and $(0.4,0.6)$, respectively. Routing heads and view projections read the same hidden state; each selected expert runs once. The balancing bias is set to zero. Shared experts, output gating, and up-projection are omitted.}
\label{fig:first_page_concept}
\end{figure}

\section{Introduction}
\label{sec:intro}

\begin{figure*}[t]
\centering
\includegraphics[width=\textwidth]{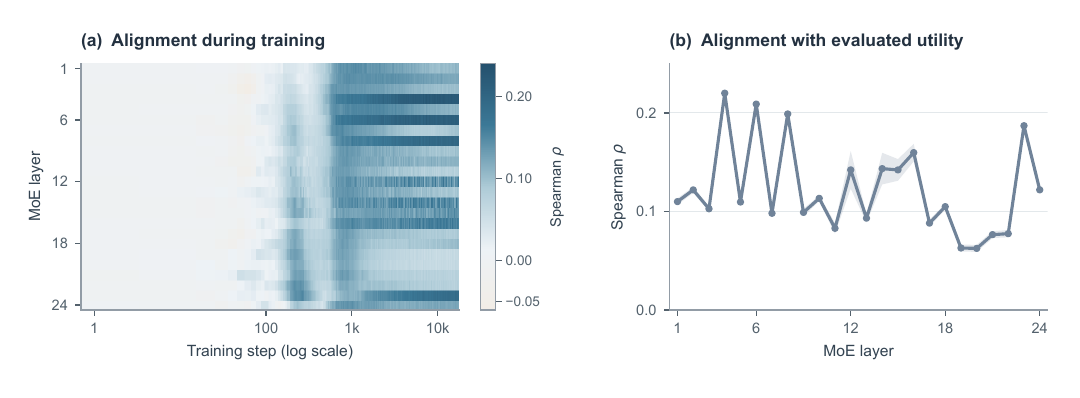}
\caption{Agreement between single-router scores and empirical expert-utility evaluation on the 24-layer 10B-A1B backbone. \textbf{(a)} Spearman correlation across training. The logarithmic training-step axis expands the early stage and compresses later intervals to show early-training changes more clearly. \textbf{(b)} Layer-wise means over the 201 logged measurements used in the router comparison; shaded bands show $\pm 1$ standard deviation. Scores exclude the load-balancing bias. Expert utility is evaluated using each expert's output $L^2$ norm, normalized by its mean over sampled tokens (App.~\ref{app:router_diagnostics}).}
\label{fig:intro_obs}
\end{figure*}

Sparse mixture-of-experts (MoE) models expand parameter capacity by activating a small subset of experts for each token \citep{shazeer2017outrageously,fedus2021switch}. This conditional computation enables larger language models under a limited active-computation budget. Routing scores determine which experts process each token and how their outputs are combined. The resulting capability depends on both expert selection and the computation performed by the selected experts.

Traditional MoE uses a single routing head followed by hard top-$k$ selection. Dense evaluation of all experts reveals limited agreement between the router's ranking and the ranking obtained from empirical expert-utility evaluation, especially in middle and deep layers (Fig.~\ref{fig:intro_obs}). Unselected experts receive no direct task-loss gradient from that token, so poor selections may become self-reinforcing. This motivates assessing experts from multiple scoring perspectives rather than relying on one router.

Standard MoE uses routing scores to select experts and weight their outputs, but does not reuse them to construct expert inputs. Load-balancing methods regulate expert utilization \citep{wang2024auxiliary}, but do not themselves provide multiple assessments of each expert. Activation-based routing such as AoE obtains expert-specific signals through partial evaluations of all experts, using low-rank factorizations to control the cost \citep{lv2025aoe}. Evaluating multiple input representations separately would also add expert computation. The challenge is therefore to obtain multiple assessments of experts and use routing preferences to modulate their computation while preserving the active-expert count, expert FLOPs, and communication volume.

We propose Multi-View Latent MoE, a framework that reuses router preferences to improve expert selection and modulate expert computation. Its Multi-Router submodule combines multiple assessments of a shared expert pool: independently initialized sigmoid routing heads produce scores that are summed before one top-$k$ selection. These routers develop highly differentiated preferences during training. Their aggregation allows experts strongly favored by one router or jointly supported by multiple routers to be selected. This submodule retains a shared expert input while enriching the information used for selection.

Multi-View further uses these differentiated preferences to modulate each selected expert's computation. The summed score alone does not retain the routers' relative preferences: score vectors $(0.4,0.1)$ and $(0.2,0.3)$ both sum to $0.5$ but express opposite preferences (Fig.~\ref{fig:first_page_concept}). Each router is paired with a learned latent projection, and each selected expert's raw sigmoid scores are normalized across routers to blend the views into its input before evaluation. The projections are shared across experts, while the mixing proportions depend on both the token and the expert. Mixing takes place before dispatch, and each selected expert performs one forward computation, preserving the active-expert count, expert FLOPs, and all-to-all communication volume.

On a 24-layer 10B-A1B model trained for 708B tokens, the full Multi-View Latent MoE improves the nine-task downstream mean by $1.18$ percentage points over the baseline, reaching $34.25$; its Multi-Router submodule alone provides a $0.6$-point gain. Routing diagnostics show that this submodule improves global score--utility rank correlation by a relative $8.1\%$, with a relative gain of approximately $58\%$ in the layer with the lowest single-router alignment (\S\ref{sec:utility}). Measurements show distinct expert inputs for the same token. At the final MoE layer, permuting view-mixing proportions across experts while fixing expert selection and output gates raises held-out NLL by $0.01340$ nats/token; permutation across tokens for the same expert also increases NLL (\S\ref{sec:view}). These results show that the trained model uses the correspondence between mixing proportions, experts, and tokens.

\paragraph{Contributions.}
\begin{itemize}
\itemsep0.15em
\item \textbf{Multi-Router for improved expert ranking.} Dense evaluation of all experts reveals weak agreement between single-router score rankings and rankings from expert-utility evaluations, especially in middle and deep layers. Motivated by this observation, the Multi-Router submodule aggregates sigmoid scores from independently initialized routers. These routers develop highly differentiated expert preferences during training, and their aggregated scores achieve closer agreement with the evaluated utility rankings.
\item \textbf{Multi-View Latent MoE at a fixed expert budget.} The framework couples its Multi-Router selection submodule with router-conditioned latent-view mixing. Mixing before dispatch retains one forward computation per selected expert, the active-expert count, expert FLOPs, and all-to-all communication volume.
\item \textbf{Quantitative gains.} On a 24-layer 10B-A1B model trained for 708B tokens, full Multi-View Latent MoE improves the nine-task downstream mean by $1.18$ percentage points over the baseline; Multi-Router alone provides a $0.6$-point gain. Multi-Router improves global score--utility rank correlation by a relative $8.1\%$, with a relative gain of approximately $58\%$ in the layer with the lowest single-router alignment.
\end{itemize}

\section{Related Work}

\label{sec:related}

\paragraph{Sparse MoE and expert specialization.}

Sparse MoE layers route each token to a small subset of experts and combine their outputs \citep{shazeer2017outrageously}. Token routing with a linear routing head and top-$k$ selection remains common in sparse MoE models \citep{cai2024survey,deepseekai2024deepseekb,jiang2024mixtral}. Switch Transformer uses top-1 routing \citep{fedus2021switch}. Recent large language models also adopt MoE architectures \citep{glmteam2026glm,wang2026ernie}. Approaches such as fine-grained experts with an always-active shared expert increase the flexibility of expert specialization \citep{dai2024deepseekmoe}. In these architectures, however, a discrete routing decision determines the active experts, which process a shared token representation.

\paragraph{Router design.}

Router design addresses routing quality, stability, and load balance through methods including router $z$-loss \citep{zoph2022st}, expert-choice routing \citep{zhou2022mixture}, auxiliary-loss-free balancing \citep{wang2024auxiliary}, mitigation of representation collapse \citep{chi2022representation}, and attention over experts \citep{wu2024yuan}. Expert assignments can fluctuate during training, motivating methods for more stable routing \citep{dai2022stablemoe,nguyen2025improving}. Router Upcycling initializes query projections and expert keys from preceding attention layers for dense-to-MoE upcycling, then sums query--key matching scores across routers before softmax selection \citep{ran2025router}. MoR uses a learnable main router to weight sub-router outputs in LoRA expert mixtures for parameter-efficient fine-tuning \citep{zhang2025mor}. Our Multi-Router submodule sums scores from separately parameterized sigmoid routing heads during pretraining. Multi-View further uses each expert's relative scores across routing heads to construct its input.

\paragraph{Multi-representation and latent-space MoE.}

A separate line of work changes the representations processed by experts. LatentMoE computes router scores from the full hidden state and evaluates experts in a low-dimensional latent space \citep{elango2026latentmoe}. Stable LatentMoE modifies this design for training stability while retaining a single down-projection \citep{team2026kimi}. We build on this shared-projection architecture. MoLAE factorizes expert weights into shared latent mappings, optionally within expert groups, and expert-specific transformations \citep{liu2025molae}. AoE removes the separate router: experts are ranked by the $L^2$ norms of their low-dimensional internal activations, and only selected experts continue computation; low-rank factorization reduces the cost of this initial evaluation \citep{lv2025aoe}. MH-MoE splits projected tokens into sub-tokens that share a router and expert pool \citep{wu2024multi,huang2024mh}. Multi-Head LatentMoE instead assigns each sub-token an independent MoE module with its own router and experts \citep{cui2026multi}; SliceMoE also explores token slicing \citep{vejendla2025slicemoe}. MoH applies shared and routed components to attention heads rather than to the FFN \citep{jin2024moh}.

\method{} uses $K$ projections of the full hidden state as a shared basis. For each selected expert, raw scores are normalized across routers to determine a token- and expert-dependent mixture before expert computation. The number of projections depends on $K$, rather than on the number of experts, and expert parameters are shared across views. This changes the representation each expert receives. Prior analyses relate hidden-state geometry to expert-usage patterns \citep{wang2026myth}. Multi-Router retains the shared latent input of LatentMoE while aggregating multiple routing heads; Multi-View further reuses their preferences to modulate expert inputs.

\section{Method}
\label{sec:method}

\subsection{Preliminaries and framework overview}
\label{sec:prelim}

We study autoregressive language modeling and modify the routing and expert-input interface of a latent-space MoE backbone \citep{elango2026latentmoe}. The router reads the hidden state $h\in\mathbb R^D$ and selects $k$ of $N$ routed experts. A down-projection $P\in\mathbb R^{d\times D}$ supplies the shared input $z=Ph$. Each expert maps $\mathbb R^d$ to $\mathbb R^d$. With selected set $\mathcal S$ and output gates $g_i$, the latent expert aggregation is
\begin{equation}
y=\sum_{i\in\mathcal S}g_i E_i(z).
\label{eq:baseline}
\end{equation}
The backbone up-projects this aggregation to dimension $D$ and combines it with the always-active shared-expert output.

\begin{figure*}[t]
\centering
\includegraphics[width=\textwidth]{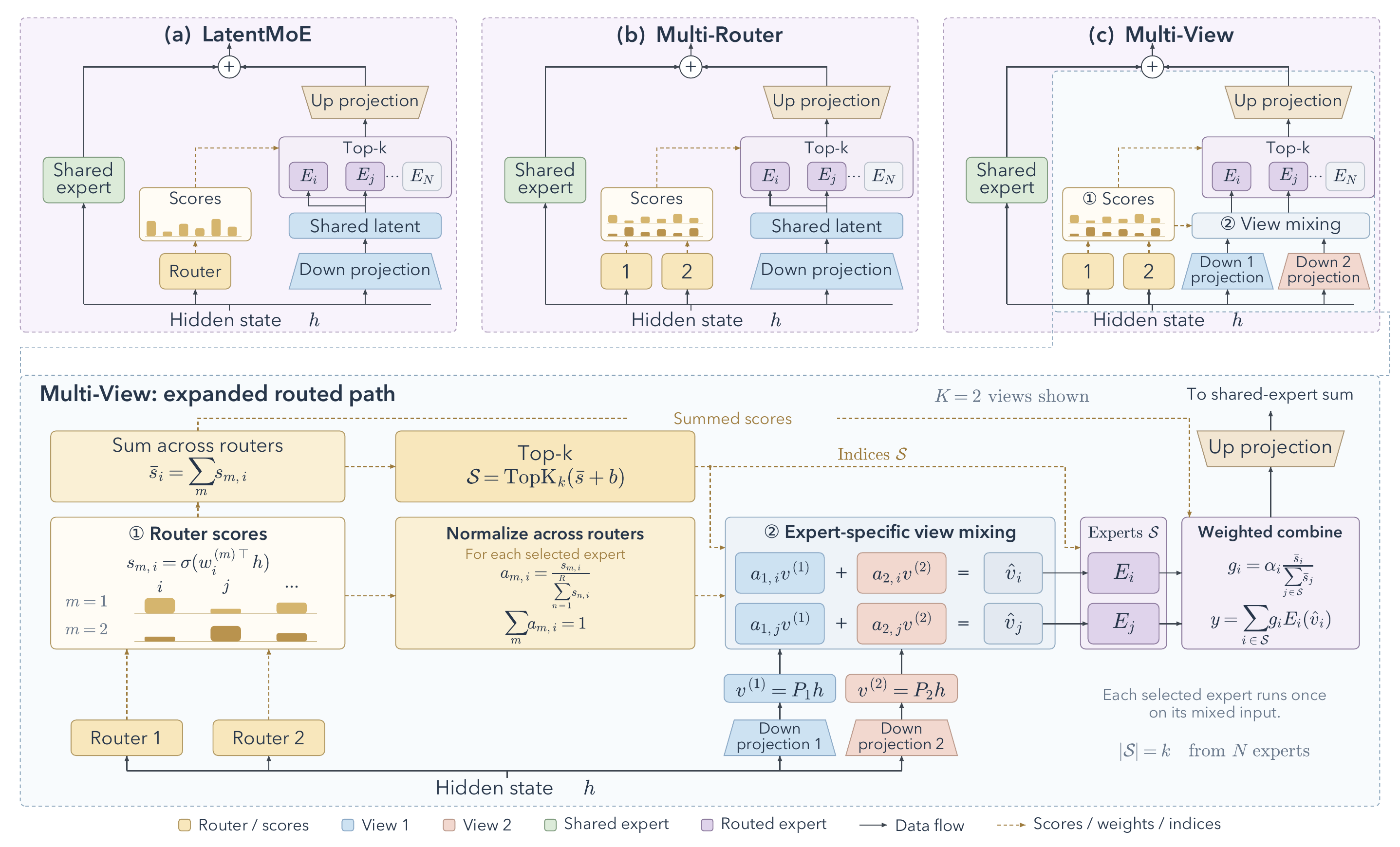}
\caption{Comparison of latent-space MoE architectures. Top: (a) LatentMoE uses a router and a shared latent input; (b) Multi-Router sums scores from separately parameterized sigmoid routing heads for top-$k$ selection while retaining the shared input; (c) Multi-View adds distinct down-projections and expert-specific view mixing. Bottom: expansion of the boxed routed path in (c), with matching colors and numbered markers. Balancing bias $b$ is added to summed scores only for selection. Output gates $g_i$ follow Eq.~\ref{eq:gate}. For each selected expert, raw sigmoid scores are normalized across routers to form view proportions $a_{m,i}$, whose sum is one. Each selected expert processes its mixed input once. The aggregated representation $y$ is up-projected and combined with the shared-expert output shown above. $K=2$ views and selected experts $i,j$ are illustrative; $k$ is the number of active experts out of $N$. Bars depict example routing scores. Normalization modules and residual paths are omitted for clarity.}
\label{fig:overview}
\end{figure*}

\mv{} retains the shared expert pool and $|\mathcal S|=k$, while reusing the preferences of individual routing heads for both selection and input construction. Its \mr{} submodule aggregates scores; for each selected expert, scores normalized across routers mix latent views before one expert evaluation. Figure~\ref{fig:overview} shows the computation.

\subsection{Multi-Router: score aggregation before selection}
\label{sec:multirouter}

\mr{} uses $R$ routing heads $\{w^{(r)}\}$, $w^{(r)}\in\mathbb{R}^{D\times N}$, to score the shared expert pool from the full hidden state $h$. Each routing head's output passes through a sigmoid, and the resulting scores are summed before top-$k$ selection:
\begin{equation}
\begin{aligned}
\bar{s}_i&=\sum_{r=1}^{R}\sigma\!\big(w_i^{(r)\top}h\big),\\
\mathcal{S}&=\mathrm{TopK}_k(\bar{s}+b).
\end{aligned}
\label{eq:mr}
\end{equation}
Here $b\in\mathbb R^N$ is the backbone's expert-balancing bias, added only for selection. Output gates normalize the uncorrected score sums over the selected experts and apply the backbone's routed-output scale:
\begin{equation}
g_i=\alpha_i\frac{\bar{s}_i}{\sum_{j\in\mathcal S}\bar{s}_j},\qquad i\in\mathcal S.
\label{eq:gate}
\end{equation}
Here $\alpha_i$ is the backbone's learnable scalar for expert $i$; it scales the expert output, not the view mixture.

Routing heads are initialized with distinct seeds to break symmetry, without a diversity penalty. We use $R=2$ by default. In \mv{} (\S\ref{sec:multiview}), each router is paired with one view, so $R=K$ and the two symbols refer to the same count.

Summation after the sigmoid preserves the routing heads as distinct nonlinear scoring functions. Summing logits before the sigmoid instead gives a single linear scorer, since $\sum_r w_i^{(r)\top}h=(\sum_r w_i^{(r)})^\top h$. Post-sigmoid summation lets a strong score from one routing head or support across routing heads affect the shared ranking and supplies the relative contributions used for input mixing. Alternatives and their properties appear in Apps.~\ref{app:fusion}--\ref{app:invariant}, with training comparisons in \S\ref{sec:abl}.

\subsection{Multi-View: router-conditioned expert inputs}
\label{sec:multiview}

\mv{} pairs $R$ routing heads with $K$ latent views in a one-to-one correspondence, so $R=K$. Selection and output gating follow the summed-score rules of \mr{}, while each selected expert's scores normalized across routers construct its input $\hat{v}_i=\sum_m a_{m,i}v^{(m)}$. Views are fused before expert evaluation, and each selected expert runs once. This parameterization avoids evaluating experts separately for each view; for nonlinear $E_i$, it is generally not equivalent to mixing outputs after separate evaluations:
\[
\begin{aligned}
&E_i\!\Big(\sum_m a_{m,i}v^{(m)}\Big)\\
&\qquad\neq\sum_m a_{m,i}E_i(v^{(m)}).
\end{aligned}
\]

\paragraph{Views.}
We form $K$ latent views $v^{(m)}=P_m h\in\mathbb{R}^{d}$, with each $P_m\in\mathbb{R}^{d\times D}$ matching the shape of the backbone down-projection. View $1$ uses the backbone's projection $P$; the remaining $K{-}1$ views use independently initialized perturbations, as described below. Each view projects the full hidden state rather than a subset of its channels. Expert architectures and input dimensions are unchanged.

\paragraph{Routing, mixing, and gating.}
All $R$ routing heads read $h$. Routing head $m$ produces $s_{m,i}=\sigma(w_i^{(m)\top}h)$ for every expert $i$, and these scores determine three operations.
\emph{(i) Selection:} sum across routers, $\bar{s}_i=\sum_m s_{m,i}$, add the balancing bias, and take one top-$k$, $\mathcal{S}=\mathrm{TopK}_k(\bar{s}+b)$, as in \mr.
\emph{(ii) View mixing:} for each selected expert $i$, normalize its raw sigmoid scores across the $R$ routing heads:
\begin{equation}
a_{m,i}=\frac{s_{m,i}}{\sum_{n=1}^{R}s_{n,i}},\qquad
\sum_{m=1}^{K}a_{m,i}=1,\quad i\in\mathcal S.
\label{eq:mixing}
\end{equation}
These proportions form the convex mixture $\hat{v}_i=\sum_m a_{m,i}v^{(m)}$. Neither the balancing bias nor the routed-output scale enters this mixture.
\emph{(iii) Output gating:} use $g_i$ from Eq.~\ref{eq:gate} to combine the selected experts' outputs. Thus, output gating normalizes summed scores over experts, whereas input mixing normalizes each expert's scores across routers. The resulting expert aggregation is
\begin{equation}
y=\sum_{i\in\mathcal{S}} g_i\, E_i\!\Big(\sum_m a_{m,i}\,v^{(m)}\Big).
\label{eq:mv}
\end{equation}
Equivalently, the input uses a token- and expert-dependent effective projection,
\[
P_{\mathrm{eff},i}(h)=\sum_m a_{m,i}(h)P_m,\qquad
\hat v_i=P_{\mathrm{eff},i}(h)h.
\]
The projection basis $\{P_m\}$ is shared across experts; only its mixing coefficients depend on $h$ and $i$. The implementation mixes the projected vectors, rather than materializing a separate matrix $P_{\mathrm{eff},i}$ for every token--expert pair.

Normalizing each expert's scores across routers to derive the mixing weights requires no additional mixing network and makes view $m$'s contribution to expert $i$'s input depend on the corresponding routing preference. Algorithm~\ref{alg:mv} gives the forward pass. Learned mixing weights and gradient coupling are evaluated in the corresponding ablation configurations in Section~\ref{sec:abl}.

\begin{algorithm}[t]
\caption{\mv{} routed-expert branch forward pass}
\label{alg:mv}
\begin{algorithmic}[1]
\Require hidden state $h$; projections $\{P_m\}_{m=1}^{K}$; routing heads $\{w^{(m)}\}_{m=1}^{R}$; experts $\{E_i\}_{i=1}^{N}$; active count $k$; balancing bias $b$; routed-output scales $\{\alpha_i\}_{i=1}^{N}$
\State $v^{(m)} \gets P_m h$ \textbf{for } $m=1,\dots,K$ \Comment{$K$ latent views}
\State $s_{m,i} \gets \sigma(w_i^{(m)\top} h)$ \Comment{per-router scores on full $h$}
\State $\bar{s}_i \gets \sum_{m} s_{m,i}$ \Comment{summed score per expert}
\State $\mathcal{S} \gets \mathrm{TopK}_k(\bar{s}+b)$ \Comment{bias only for selection}
\For{$i \in \mathcal{S}$}
  \State $a_{m,i} \gets s_{m,i}/\sum_{n=1}^{R}s_{n,i}$ \Comment{normalize across routers}
  \State $\hat{v}_i \gets \sum_{m} a_{m,i}\, v^{(m)}$ \Comment{per-expert view blend}
  \State $g_i \gets \alpha_i\bar{s}_i / \sum_{j\in\mathcal{S}} \bar{s}_j$ \Comment{output gate, Eq.~\ref{eq:gate}}
\EndFor
\State \Return $y = \sum_{i\in\mathcal{S}} g_i\, E_i(\hat{v}_i)$
\end{algorithmic}
\end{algorithm}

\paragraph{Expert-dependent mixing weights.}
The index $i$ in $a_{m,i}$ allows input modulation to vary across experts. If all selected experts for the same token share $a_{m,i}=a_m(h)$, they receive the same input $\big(\sum_m a_m(h)P_m\big)h$. This remains token-conditioned modulation, but no longer distinguishes different experts for that token. If the weights are further fixed to constants, such as $a_m=1/K$, the mixture can be absorbed into a fixed projection. \mv{} derives $a_{m,i}$ from the raw scores for expert $i$, normalized across the $R$ routing heads, making modulation depend on both the token and the expert (App.~\ref{app:props}).

\paragraph{Initialization.}
To introduce the additional projections gradually, we retain the backbone down-projection for view~$1$ and parameterize the others as ramped perturbations, $v^{(m)}=Ph+\gamma(t)\,\Delta P_mh$, with $\gamma$ increasing from $0$ to $1$ early in training. At $t=0$, all view projections equal the backbone projection; after the ramp, the view projections are unconstrained. The expert inputs are weighted combinations of these views. This procedure can also initialize the views from an existing checkpoint (App.~\ref{app:warm}).

\paragraph{Shared expert parameters.}
All views use one expert pool, so expert parameters are not duplicated or partitioned across views. Each selected expert processes one mixed input, and gradients propagate through every view with nonzero mixing weight. \mr{} uses the shared input $z=Ph$ for all selected experts; \mv{} adds input modulation through the router-derived coefficients (App.~\ref{app:props}).

\section{Experiments}
\label{sec:exp}

\subsection{Setup}
\label{sec:setup}
We compare the latent-space MoE baseline, \mr{}, and score-derived \mv{} on a 24-layer 10B-A1B backbone with top-8 routing and one always-active shared expert \citep{elango2026latentmoe}. The three configurations share the training data, token budget, backbone, and optimization settings (App.~\ref{app:arch10b}). Downstream evaluation uses the 708B-token checkpoints on nine tasks (protocol in App.~\ref{app:more}). We use $R$ for the number of routing heads and $K$ for the number of views. The main method \mv{}($K$) pairs them one to one, so $R=K$; \mr{}($R$) uses $R$ routing heads with a shared expert input.

Training ablations examine router aggregation, learned input-mixing weights, and view-fed scoring. Their losses are averaged over a $\pm1\%$ token window around a common ablation checkpoint. A pair of near-matched runs differs after smoothing by at most $1.2\times10^{-3}$, with a mean absolute difference of $7\times10^{-4}$. Architecture, optimization, and comparison settings are detailed in App.~\ref{app:repro}.

\subsection{Downstream performance}
\label{sec:main}
\begin{table*}[t]
\centering
\small
\setlength{\tabcolsep}{3.5pt}
\renewcommand{\arraystretch}{1.08}
\resizebox{\textwidth}{!}{%
\begin{tabular}{l cccc>{\columncolor{gray!15}}c ccc>{\columncolor{gray!15}}c >{\columncolor{gray!15}}c >{\columncolor{gray!15}}c >{\columncolor{gray!30}}c}
\toprule
 & \multicolumn{5}{c}{Exams} & \multicolumn{4}{c}{Knowledge} & \multicolumn{1}{c}{Science} & \multicolumn{1}{c}{Reasoning} & \multicolumn{1}{c}{} \\
\cmidrule(lr){2-6}\cmidrule(lr){7-10}\cmidrule(lr){11-11}\cmidrule(lr){12-12}
Method & CMMLU & MMLU & Redux & Pro & Avg. & HotpotQA & SimpleQA & PreciseWiki & Avg. & GPQA & BBH & Overall \\
\midrule
Baseline (LatentMoE) & 53.71 & 57.48 & 59.03 & 27.28 & 49.38 & 22.00 & 3.35 & 8.64 & 11.33 & 30.00 & 36.10 & 33.07 \\
\rowcolor{orange!12} \mr{} ($R{=}2$) & 53.77 & 57.50 & 58.90 & \textbf{27.71} & 49.47 & 22.13 & \textbf{3.85} & \textbf{9.16} & \textbf{11.71} & 32.86 & \textbf{37.31} & 33.69 \\
\rowcolor{blue!10} \mv{} ($K{=}2$) & \textbf{55.20} & \textbf{58.67} & \textbf{60.93} & 27.12 & \textbf{50.48} & \textbf{22.54} & 3.41 & 9.02 & 11.66 & \textbf{35.60} & 35.76 & \textbf{34.25} \\
\midrule
$\Delta$ (MV $-$ Baseline) & $+1.49$ & $+1.19$ & $+1.90$ & $-0.16$ & $+1.10$ & $+0.54$ & $+0.06$ & $+0.38$ & $+0.33$ & $+5.60$ & $-0.34$ & $+1.18$ \\
\bottomrule
\end{tabular}}
\caption{Downstream scores (\%) at 708B training tokens. Baseline: latent-space MoE; MR: Multi-Router ($R=2$); MV: score-derived Multi-View ($K=R=2$; Eq.~\ref{eq:mv}). Tasks are grouped into exams, knowledge, science, and reasoning. Avg.\ is the unweighted within-group mean; Overall is the unweighted mean across all nine tasks. Bold marks the highest displayed score in each column. The final row reports MV minus Baseline in percentage points. Prompt and scoring protocols appear in App.~\ref{app:more}.}
\label{tab:bench}
\end{table*}

Table~\ref{tab:bench} compares all three configurations at the same training budget. The nine-task means are $33.07$ for the baseline, $33.69$ for \mr{}, and $34.25$ for \mv{}. Multi-View improves the mean by $1.18$ percentage points over the baseline and by $0.56$ over Multi-Router. Gains over Multi-Router reach $2.74$ points on GPQA and $2.03$ on MMLU-Redux; CMMLU and MMLU improve by $1.43$ and $1.17$ points, respectively. These results indicate an aggregate benefit from extending selection with multiple routing heads to router-conditioned expert computation.

\begin{figure*}[t]
\centering
\includegraphics[width=\textwidth]{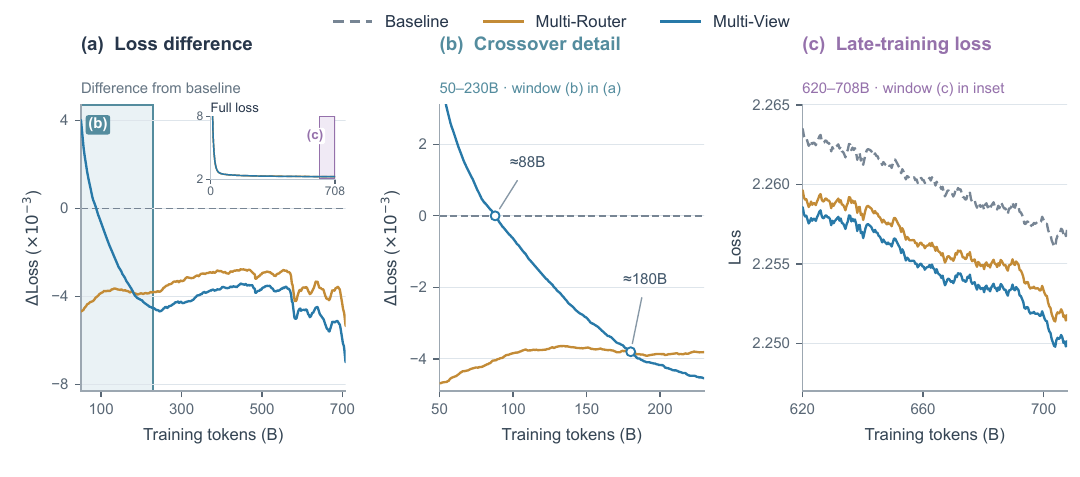}
\caption{Training loss of the baseline, Multi-Router, and Multi-View on 10B-A1B through 708B tokens. \textbf{(a)} Loss difference from the baseline; the inset shows the full loss trajectories. \textbf{(b)} Magnification of the blue-bordered window in (a), covering 50--230B tokens. \textbf{(c)} Magnification of the purple-bordered window in the full-loss inset, showing absolute loss over 620--708B tokens. All curves use the same causal exponential moving average (EMA), with a 2,000-step span ($\alpha=2/2001$), initialized at the first observation. The final losses reported in the text average the last 1,000 unsmoothed steps.}
\label{fig:loss}
\end{figure*}

\paragraph{Training convergence.}
Figure~\ref{fig:loss} compares the three configurations through the 708B-token checkpoint used for downstream evaluation. Over the final 1,000 steps, the mean losses are $2.25621$ for the baseline, $2.25047$ for \mr{}, and $2.24883$ for \mv{}. Multi-Router reduces loss by $5.74\times10^{-3}$ relative to the baseline, and Multi-View increases this reduction to $7.38\times10^{-3}$. Multi-View's lower loss in late training is consistent with its downstream gains.

\subsection{Router preferences shape expert computation}
\label{sec:view}
\begin{figure*}[t]
\centering
\includegraphics[width=\textwidth]{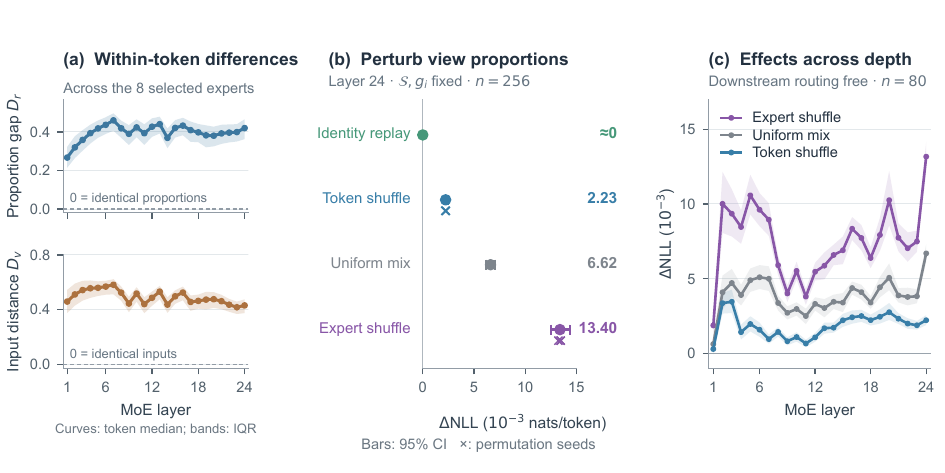}
\caption{Expert-specific inputs and controlled mixing interventions in the score-derived \mv{} checkpoint at 708B tokens. \textbf{(a)} Differences among the eight experts selected for the same token: the mean absolute pairwise difference in view-1 proportions $D_r$ (top) and the mean pairwise input distance normalized by the mean input norm $D_v$ (bottom). Zero denotes identical proportions or inputs, respectively. Curves show the median across tokens at each layer; bands show the 25th--75th percentiles. \textbf{(b)} Final-layer interventions on 256 sequences, changing view proportions while holding $\mathcal{S}$ and output gates $g_i$ fixed to numerical precision. Whiskers are 95\% sequence-bootstrap CIs; crosses mark permutation-seed means. \textbf{(c)} Single-layer interventions on 80 sequences. Bands are pointwise 95\% CIs; downstream routing is recomputed. $\Delta$NLL is intervention loss minus the original loss.}
\label{fig:mv_mechanism}
\end{figure*}

We probe the score-derived \mv{} checkpoint used for downstream evaluation at 708B tokens. The analysis covers all 24 MoE layers on 256 held-out sequences of 4096 tokens. It examines expert-specific inputs and whether the trained model uses the correspondence between mixing proportions, experts, and tokens.

\paragraph{From preferences to distinct expert inputs.}
The two routing heads have markedly different rankings: across layers, their top-8 Jaccard index averages $0.145$ and their top-1 agreement averages $11.2\%$. For each token, the probe exports the actual view-1 mixing proportion $r_{1,i}=a_{1,i}$ and mixed input of each of its eight selected experts. We compare all 28 pairs of these experts: $D_r$ is their mean absolute proportion difference, and $D_v$ is their mean input distance divided by their mean input norm (App.~\ref{app:mv_probes}). Both are zero when all eight experts receive the same proportions or inputs, respectively. Fig.~\ref{fig:mv_mechanism}(a) summarizes these within-token differences across tokens at each layer. The layer-wise medians are $0.27$--$0.46$ for $D_r$ and $0.42$--$0.58$ for $D_v$. The selected experts therefore receive distinct inputs according to their view preferences.

\paragraph{The trained model uses the correspondence.}
At the final MoE layer, we hold the selected experts $\mathcal{S}$ and their output gates $g_i$ fixed to numerical precision and intervene only on the view-mixing proportions. For a replacement proportion $r_{1,i}^{\prime}$, the two dispatcher channels are rebuilt as $(g_i r_{1,i}^{\prime},g_i(1-r_{1,i}^{\prime}))$, preserving their sum $g_i$. Permuting the view proportions among the selected experts of the same token raises NLL by $0.01340$ nats/token (95\% sequence-bootstrap CI: $[0.01256,0.01435]$). Using uniform mixing proportions $(1/2,1/2)$ raises NLL by $0.00662$; permuting proportions across tokens for the same expert raises it by $0.00223$. The respective CIs are $[0.00620,0.00709]$ and $[0.00205,0.00241]$. Expert permutation increases sequence-average NLL for all 256 sequences. These interventions show that the trained checkpoint uses the correspondence between mixing proportions, experts, and tokens. Expert permutation produces the largest increase among these three interventions.

\paragraph{Both views contribute across depth.}
Applying the view~1-only or view~2-only proportions at the final layer increases NLL by $0.01881$ and $0.02042$ nats/token, respectively. The trained checkpoint therefore uses both view branches. A separate 80-sequence scan perturbs one layer at a time: expert permutation and uniform mixing increase NLL, with pointwise 95\% CIs above zero at all 24 layers. Only the intervened layer's selections and output gates are fixed; subsequent routing is recomputed, so these are total effects including downstream responses. Additional preference, single-view, and token-group analyses appear in App.~\ref{app:mv_probes}.

\subsection{Alignment of router scores and expert outputs}
\label{sec:utility}
\begin{figure*}[t]
\centering
\includegraphics[width=\textwidth]{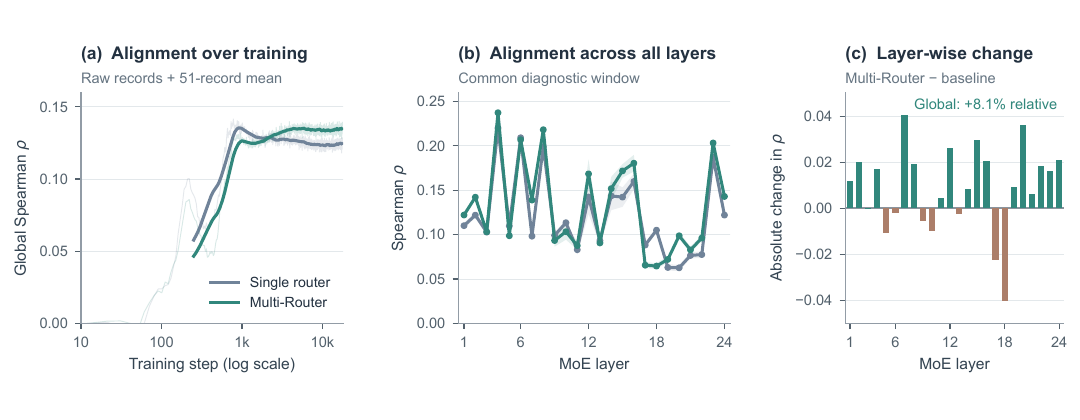}
\caption{Router-score alignment with normalized expert-output norms. \textbf{(a)} Global Spearman correlation over training; solid curves use a 51-record moving mean and pale curves show raw measurements. \textbf{(b)} Layer-wise means and one-standard-deviation bands over 201 matched logged measurements; metric and aggregation definitions appear in App.~\ref{app:router_diagnostics}. \textbf{(c)} Correlation changes across all 24 layers. Multi-Router improves global correlation by a relative $8.1\%$, with the largest relative layer-wise gain of $57.7\%$ at layer~20.}
\label{fig:oracle_align}
\end{figure*}

\begin{table}[t]
\centering\small
\setlength{\tabcolsep}{8pt}
\begin{tabular}{lc}
\toprule
Method & Spearman $\rho$\\
\midrule
Single router & $0.1245$\\
Multi-Router & $\mathbf{0.1346}$\\
\bottomrule
\end{tabular}
\caption{Global alignment between pre-bias router scores and normalized expert-output norms, averaged over 201 matched logged measurements. Multi-Router improves rank correlation by a relative $8.1\%$. Definitions appear in App.~\ref{app:router_diagnostics}.}
\label{tab:utility}
\end{table}

We compare the baseline and Multi-Router using the normalized expert-output norm. For each sampled token, all 256 experts process its latent input; each output's $L^2$ norm is divided by that expert's mean norm over the sampled tokens. Spearman correlation measures agreement between the resulting expert ranking and the router scores before balancing bias (App.~\ref{app:router_diagnostics}). Related work also examines routing quality through expert-pathway optimization, routing-weight analysis, and expert-importance measurements \citep{li2025c3po,li2025r2t2,li2025routing,chen2026banpick}.

Table~\ref{tab:utility} reports an $8.1\%$ relative improvement in global rank correlation across 24 MoE layers, from $0.1245$ to $0.1346$. Figure~\ref{fig:oracle_align} shows the training trajectories and layer-wise comparison, with gains in 17 of 24 layers. The largest relative gain occurs at layer~20, where correlation rises from $0.0625$ to $0.0986$ ($57.7\%$); layer~7 shows the largest absolute gain, from $0.0982$ to $0.1388$. These measurements show that aggregating differentiated routing heads improves alignment between router scores and normalized expert outputs, complementing the downstream and training-loss gains of Multi-Router.

\subsection{Training ablations}
\label{sec:abl}
\begin{table}[t]
\centering
\small
\setlength{\tabcolsep}{4pt}
\resizebox{\columnwidth}{!}{%
\begin{tabular}{lc}
\toprule
Method & $\Delta$ loss vs.\ baseline ($\downarrow$) \\
\midrule
Single router (baseline; loss $2.3825$) & $0.0000$ \\
\mr{} ($R{=}2$, sigmoid-sum)            & $-0.0045$ \\
\mr{} ($R{=}3$)                         & $-0.0051$ \\
\midrule
\multicolumn{2}{l}{\emph{Fusion position} ($R{=}2$)} \\
\quad Logit-sum                       & \emph{collapses to one router} \\
\quad Probability-normalized sum      & $-0.0037$ \\
\midrule
\multicolumn{2}{l}{\emph{Fusion operator} (post-sigmoid, $R{=}2$)} \\
\quad Max $\max(s_1,s_2)$             & $-0.0037$ \\
\quad Max to select, sum to gate      & $-0.0036$ \\
\quad Geometric mean $\sqrt{s_1s_2}$  & $-0.0037$ \\
\quad Product $s_1s_2$                & $-0.0034$ \\
\quad Sum $+\,\lambda s_1s_2$ (agreement)        & $-0.0037$ \\
\quad Sum $-\,\lambda|s_1{-}s_2|$ (disagreement) & $-0.0033$ \\
\midrule
Observed run-pair difference   & $\pm0.0012$ \\
\bottomrule
\end{tabular}}
\caption{Multi-Router training ablations on 10B-A1B. Loss differences use the single-router baseline and a $\pm1\%$ token window around the common ablation checkpoint. All post-sigmoid aggregation operators reduce loss; the final row reports the observed run-pair difference.}
\label{tab:main}
\end{table}

\begin{table}[t]
\centering
\small
\setlength{\tabcolsep}{4pt}
\resizebox{\columnwidth}{!}{%
\begin{tabular}{lc}
\toprule
Variant ($K{=}2$) & $\Delta$ loss vs.\ baseline ($\downarrow$) \\
\midrule
\mr{} ($R{=}2$)           & $-0.0045$ \\
\midrule
\multicolumn{2}{l}{\emph{Learned per-expert view weights}} \\
\quad Decoupled gate, init $0.5$      & $-0.0048$ \\
\quad Decoupled gate, asym.\ init     & $-0.0050$ \\
\quad Two independent gates           & $-0.0044$ \\
\midrule
\multicolumn{2}{l}{\emph{Coupling between scoring and views}} \\
\quad Coupled (gradient flows to views) & $-0.0024$ \\
\quad Detached (gradient stopped)       & $+0.0090$ \\
\midrule
Observed run-pair difference   & $\pm0.0012$ \\
\bottomrule
\end{tabular}}
\caption{Input-mixing training ablations on 10B-A1B. The learned-weight variants use per-expert mixing parameters. In the view-fed coupling pair, both scoring and expert computation read the views; stopping the weight-to-view gradient increases loss by $0.0114$ while preserving forward weights at fixed parameters.}
\label{tab:mv_abl}
\end{table}

\paragraph{Fusion position and routing heads.}
Multi-Router reduces loss by $4.5\times10^{-3}$ with two routing heads and $5.1\times10^{-3}$ with three routing heads (Table~\ref{tab:main}). The additional reduction of $0.6\times10^{-3}$ is smaller than the observed run-pair difference of $1.2\times10^{-3}$. We use two routing heads by default. Fusion position determines whether the routing heads contribute distinct scoring functions: summing logits before sigmoid collapses them into one linear scorer (\S\ref{sec:multirouter}), whereas applying sigmoid to each routing head before summation preserves their nonlinear contributions. The latter distinguishes equal-mean logits with different distributions across routing heads.

\paragraph{Fusion operators and routing-head contributions.}
Different fusion rules determine how the scoring views contribute to expert selection. Maximum fusion retains only the strongest score, whereas products and geometric means emphasize joint support. Summation accumulates bounded contributions from all routing heads, allowing either a strong response from one routing head or support across routing heads to increase the aggregate. All evaluated fusion variants reduce loss by $3.3$--$4.5\times10^{-3}$ (Fig.~\ref{fig:operators}). The operator gaps lie within the observed run-pair difference. Beyond selection, \mv{} normalizes each selected expert's raw scores across routing heads to construct input-mixing proportions (\S\ref{sec:multiview}), connecting selection with multiple routing heads to router-conditioned expert computation.

\paragraph{Learned input mixing.}
Table~\ref{tab:mv_abl} compares three learned per-expert mixing parameterizations: symmetric initialization at $0.5$, asymmetric initialization, and two independent gates. They reduce loss by $4.4$--$5.0\times10^{-3}$ relative to the baseline while retaining the same active-expert budget. The main \mv{} method instead derives mixing weights directly from router scores, reusing the selection signal without an additional mixing network.

\paragraph{Gradient coupling in view-fed scoring.}
A separate ablation feeds the views into both the scoring and expert-computation paths. Stopping the gradient from mixing weights to views changes the loss difference from $-0.0024$ to $+0.0090$, an increase of $0.0114$. The forward weight values at fixed parameters are preserved, isolating the benefit of this gradient path within the view-fed configuration.

\subsection{Parameter and computational efficiency}
\label{sec:cost}
\paragraph{Parameter overhead.}
For the 24-layer 10B-A1B backbone with two routing heads, the Multi-Router submodule adds approximately \paramDeltaMR{} parameters. Multi-View adds one latent-view projection per layer, contributing a further $0.13\%$, for a total increase of approximately \paramDeltaMV{}.

\paragraph{Expert computation and communication.}
Each selected expert processes one mixed input, preserving the active-expert count and activated expert FLOPs. Input mixing takes place on the source device before dispatch; each token still sends $k$ vectors of dimension $d$, preserving all-to-all communication volume. Additional work comes from routing, projection, and input mixing; for fixed $K$, the extra scoring and projection cost scales as $D(d+N)$. Appendix~\ref{app:cost} gives the parameter and operation counts, together with same-cluster throughput measurements for the router and view-fed ablation configurations.

\section{Conclusion}
\label{sec:conclusion}

Multi-View Latent MoE reuses routing preferences to improve expert selection and shape expert computation. Motivated by weak agreement between single-router rankings and empirical expert-utility rankings, its Multi-Router submodule combines independently initialized routing heads that learn highly differentiated preferences. Routing diagnostics show relative rank-correlation gains of $8.1\%$ globally and approximately $58\%$ in the layer with the lowest single-router alignment. Multi-View further uses each selected expert's relative scores across routing heads to blend latent views into its input before dispatch. Each selected expert runs once, preserving the active-expert count, expert FLOPs, and all-to-all communication volume. On a 24-layer 10B-A1B model trained for 708B tokens, full Multi-View Latent MoE improves the nine-task downstream mean by $1.18$ percentage points over the baseline; its Multi-Router submodule alone provides a $0.6$-point gain. Controlled mixing interventions show that the model makes effective use of expert- and token-dependent input differences. These findings establish routing preferences as a signal for both selecting experts and shaping how they process each token.

\section*{Limitations}
\label{sec:limitations}

The experiments use a 24-layer 10B-A1B backbone with a shared training seed and evaluate nine downstream tasks at the 708B-token checkpoint. The controlled interventions measure the trained model's use of view mixing at that checkpoint.

\bibliography{references}

\clearpage
\appendix

\section{Extended Method Discussion}
\label{app:method}

This appendix provides supporting analyses for the main-text method. Algorithm~\ref{alg:mv} in \S\ref{sec:multiview} gives the forward pass of the \mv{} routed-expert branch.

\subsection{LatentMoE as the backbone}
\label{app:backbone}
Because the experts operate in latent space, replacing a selected expert's input $z$ with a view $v^{(m)}$ leaves its architecture, width, and hyperparameters unchanged. The existing expert blocks can therefore be reused without increasing expert capacity. Latent-space designs have also been studied in other MoE architectures \citep{elango2026latentmoe,liu2025molae}.

\subsection{Contextual representations and expert inputs}
\label{app:attention}
Attention produces a contextual representation $h$, but the baseline MoE still evaluates all $k$ selected experts at the same projected input $z=Ph$. The attention representation and the expert inputs are separate architectural choices: multiple projections can give the selected experts different inputs derived from the same $h$. We use the term \emph{view} for this learned latent projection or the resulting expert input.

\subsection{Choice of fusion position}
\label{app:fusion}
We consider three positions for combining $R$ router scores. \emph{(1) Before the nonlinearity (logit sum):} $\sum_r w_i^{(r)\top}h=(\sum_r w_i^{(r)})^{\top}h$ is equivalent to a single scorer with weights $\sum_r w_i^{(r)}$ (\S\ref{sec:multirouter}). \emph{(2) After the sigmoid (ours):} each router first applies its own nonlinearity, after which the resulting scores are combined. \emph{(3) After normalization into probabilities:} each router's scores are normalized before summation, removing differences in their overall scale. Table~\ref{tab:main} reports the corresponding training comparisons.

\subsection{Choice of fusion operator}
\label{app:operator}
For post-sigmoid fusion, different aggregation operators emphasize different score patterns. Let $s_m\in(0,1)$ denote one router's score for an expert. An additive operator, $\sum_m s_m$, allows a high score from one router to increase the aggregate even when other scores are low. Multiplicative operators, $\prod_m s_m$ and the geometric mean $(\prod_m s_m)^{1/R}$, suppress an expert when any router assigns it a near-zero score. We also consider additive operators with an agreement term, $\sum_m s_m + \lambda\prod_m s_m$, or a disagreement penalty, $\sum_m s_m - \lambda\,|s_1-s_2|$; both reduce to summation at $\lambda=0$.

Summation has two useful properties, with empirical comparisons in Table~\ref{tab:main} and Fig.~\ref{fig:operators}. First, multiplicative aggregation is sensitive to low scores: one near-zero score can suppress an expert regardless of the other routers' scores. Summation does not impose this requirement for agreement. Second, with $\bar{s}_i=\sum_m s_{m,i}$, the expert's output gate in Eq.~\ref{eq:gate} can be decomposed into per-router contributions, $g_i=\sum_m \alpha_i s_{m,i}/\sum_{j\in\mathcal S}\bar{s}_j$. A product does not have this additive decomposition, and its scale depends on $R$. The same raw sigmoid scores also supply the input-mixing weights in \S\ref{sec:multiview}: for each selected expert, its scores are normalized across routers, $a_{m,i}=s_{m,i}/\sum_n s_{n,i}$, so that $\sum_m a_{m,i}=1$.

We also evaluate a variant that selects experts using $\max_m s_{m,i}$ and gates their outputs using $\sum_m s_{m,i}$. Selection therefore depends on an expert's highest router score, while the output weight uses scores from all routers. This separates the effects of score aggregation on selection and gating. Conceptually, it corresponds to choosing one router at the selection stage, analogous to choosing one view for an expert's input in a hard single-view variant. Table~\ref{tab:mv_abl} separately reports the evaluated view-mixing variants.

\subsection{Aggregation before selection}
\label{app:invariant}
The layer aggregates all router scores before a single top-$k$ selection. Every router scores the full expert set $\{1,\dots,N\}$, and the number of activated experts remains $k$, independent of the number $R$ of routers or $K$ of views. All routing heads contribute to one shared ranking over the full expert pool.

An alternative is to distribute a fixed budget of $k$ selections adaptively among routers. Although this preserves the total budget, each router still ranks experts independently and selects only up to its assigned quota. An expert may therefore be excluded from each local selection even if its aggregate score would place it in the global top-$k$. We instead aggregate the scores and perform one selection with a fixed total of $k$ experts.

\subsection{Cost accounting in detail}
\label{app:cost}
Let $L$ be the number of MoE layers. The hidden dimension is $D$, the latent dimension is $d$, and there are $N$ routed experts with $k$ active per token. For the main configuration with $R=K$, each layer adds $K-1$ projections of shape $d\times D$ and $R-1$ routing heads of shape $D\times N$. Thus,
\begin{equation}
\begin{aligned}
\Delta\theta_{\mathrm{layer}}&=(K-1)D(d+N),\\
\Delta\theta_{\mathrm{model}}&=L\Delta\theta_{\mathrm{layer}}.
\end{aligned}
\label{eq:params}
\end{equation}
With $L=24$, $K=2$, $D=1024$, $d=512$, and $N=256$, \mv{} adds $786{,}432$ parameters per layer and $18{,}874{,}368$ in total, approximately $0.19\%$ of the 10B-parameter backbone. \mr{} adds only the routing heads: $262{,}144$ parameters per layer and $6{,}291{,}456$ in total, approximately $0.06\%$.

The expert path still evaluates $k$ unchanged experts on $d$-dimensional inputs. For each token and layer, the additional projections and scoring matrices require $(K-1)D(d+N)$ multiply--accumulate operations (MACs). Counting one MAC as two FLOPs, and a direct mixture as $K$ multiplications and $K-1$ additions per coordinate, the additional forward cost for $K\ge2$ is approximately
\begin{equation}
\begin{aligned}
\Delta F_{\mathrm{forward}}\simeq{}&2(K-1)D(d+N)\\
&+kd(2K-1).
\end{aligned}
\label{eq:flops}
\end{equation}
The estimate covers scoring and projection matrix multiplications and input blending per token and layer. Expert-block FLOPs are unchanged. The implementation still dispatches $k$ vectors of dimension $d$ per token, preserving all-to-all volume (App.~\ref{app:impl}).

Table~\ref{tab:cost} reports throughput for scoring-only and coupled/detached view-fed variants on one cluster.
\begin{table}[t]
\centering\small
\setlength{\tabcolsep}{4pt}
\begin{tabular}{lrr}
\toprule
Variant & Expert FLOPs & Throughput \\
\midrule
\mr{} ($R=2$) & $1.00\times$ & $17{,}375$ \\
Geometric mean & $1.00\times$ & $17{,}424$ \\
Product & $1.00\times$ & $17{,}564$ \\
Coupled views & $1.00\times$ & $16{,}449$ \\
Detached views & $1.00\times$ & $16{,}908$ \\
\bottomrule
\end{tabular}
\caption{Same-cluster throughput in tokens/s/card for the listed 10B-A1B ablation configurations. Coupled and detached variants use view-fed scoring. Expert FLOPs are normalized to the backbone. Main-framework parameter counts are given in Eq.~\ref{eq:params}.}
\label{tab:cost}
\end{table}

\subsection{Implementation}
\label{app:impl}
Standard MoE implementations permute tokens before all-to-all communication, copying each token's latent vector once per selected expert into a buffer organized by expert. \mv{} instead constructs the $d$-dimensional mixed input $\hat{v}_i=\sum_m a_{m,i}v^{(m)}$ for each selected (token, expert) pair on the token's device before dispatch. For $T$ tokens, the implementation flattens the $T\times k$ token--expert slots, each with one selected expert as its destination. Each token still dispatches $k$ vectors of dimension $d$, so the all-to-all volume is unchanged. Dispatching all $K$ views and combining them on the expert's device would instead multiply that volume by $K$. The return all-to-all communication, inverse permutation, and gated sum of expert outputs are unchanged.

Because selection uses no per-view quota, load balancing is applied once over the $N$ experts using the backbone's existing scheme. No auxiliary loss is added for the views.

The expert bias is added after summing the routing-head scores and is used only by top-$k$ selection, $\mathcal S=\mathrm{TopK}_k(\bar s+b)$; it enters neither the input-mixing weights nor the output gates. The output gates follow Eq.~\ref{eq:gate}: the uncorrected $\bar s_i$ are normalized over the selected set and then multiplied by the backbone's per-expert output scale. For input mixing, Eq.~\ref{eq:mixing} normalizes each selected expert's raw sigmoid scores across routers, and the resulting coefficients directly form $\hat v_i=\sum_m a_{m,i}v^{(m)}$. The input-ratio computation uses $\epsilon=10^{-9}$ for numerical stability, omitted from the mathematical formulas. At fixed parameters, replacing the score sum by its mean and scaling $b$ by $1/R$ preserves selection, output gates, and view weights. To maintain the bias correspondence during training, its initialization and update increments must also be scaled by $1/R$. In Multi-Router, repeating one routing head $R$ times and multiplying its balancing bias by $R$ preserves the forward computation with a single routing head.

\paragraph{Execution-path verification.} For the checkpoint probes, we verify the expert IDs, output gates, and mixing weights received by the dispatcher. Appendix~\ref{app:mv_probes} reports these execution checks and identity-replay errors.

\subsection{Interaction with other components}
\label{app:compat}
\mv{} modifies the MoE down-projection, router, and permutation step. Attention, including multi-head latent attention (MLA) \citep{deepseekai2024deepseek}, is unchanged, as is the KV cache. Hyper-connections and their manifold-constrained variant mHC use multiple residual streams with learned mixing between sublayers \citep{zhu2024hyper,xie2025mhc}. \mv{} accepts the resulting sublayer input $h$ and returns a $D$-dimensional output, so these designs are architecturally compatible. Multi-token prediction (MTP) \citep{deepseekai2024deepseekb} adds prediction modules and a loss term, while \mv{} leaves the training objective unchanged. In our configuration, the MTP module uses a dense FFN and is not modified. The method is also compatible in design with fine-grained and shared experts, auxiliary-loss-free balancing, and dense or hash-routed early layers \citep{roller2021hash}. It leaves the vocabulary, embeddings, routed and active expert counts, expert blocks, up-projection, and expert-parallel layout unchanged.

\subsection{Warm-starting from a LatentMoE checkpoint}
\label{app:warm}
An existing LatentMoE checkpoint provides a possible initialization for \mv{}. View~1 retains the checkpoint's down-projection and routing head. The remaining views are initialized from the same projection with small perturbations, and their routing heads are initialized to zero. These additional routing heads add the same constant to every expert's raw score, preserving the raw-score ordering but generally changing normalized output gates. Keeping the balancing bias $b$ unchanged also preserves the actual bias-corrected selection, because the same constant is added to every entry of $\bar s+b$. The views can then differentiate during continued training.

\subsection{Properties: tied views and view differentiation}
\label{app:props}
\paragraph{Tied-view reduction.} Since $\sum_m a_{m,i}=1$, tying all $P_m=P$ gives $\hat v_i=(\sum_m a_{m,i})Ph=Ph$ for every selected expert. With the same router scores, selected set, and output gates, this recovers \mr{} with the shared input $z=Ph$. All routers continue to score the same expert pool, independently of whether the view projections are tied.

\paragraph{Expert-dependent mixing.} For a given token, mixing weights that vary across experts allow those experts to receive different inputs. If all selected experts share $a_{m,i}=a_m(h)$, then
\[
\begin{aligned}
\hat{v}_i&=\sum_m a_m(h)\,P_m h\\
&=\Big(\sum_m a_m(h)\,P_m\Big)h=P_{a(h)}\,h,
\end{aligned}
\]
All $k$ experts then share an input, but $P_{a(h)}$ can still vary with the token, so input modulation remains. When $a_m$ is constant, including $a_m=1/K$, the mixture can be absorbed into a fixed projection. The main method's $a_{m,i}$ retains both token and expert dependence, providing modulation at this granularity without an additional mixing network.

\paragraph{Per-view load.} Aggregate load balancing (App.~\ref{app:impl}) controls the number of tokens assigned to each expert, which affects dispatch cost and expert utilization. For the per-view diagnostics, let $u_{m,i}(h_t)$ denote the recorded weight of view $m$ at expert $i$. Define $p_{m,i}=\sum_{t:\,i\in\mathcal{S}(h_t)}u_{m,i}(h_t)\big/\sum_{j}\sum_{t:\,j\in\mathcal{S}(h_t)}u_{m,j}(h_t)$ as the fraction of view $m$'s recorded weight mass assigned to expert $i$. Aggregate assignment counts can remain unchanged while $p_{m,\cdot}$ varies. Different views can therefore concentrate on different subsets even when the aggregate expert load is uniform. We report the coefficient of variation of $p_{m,\cdot}$ and the effective expert count $\exp(H_m)$, where $H_m$ is the entropy of $p_{m,\cdot}$ (App.~\ref{app:view_load}). A low effective count indicates concentration on fewer experts. We also report the ratio of an expert's selection share to its gate-mass share, which distinguishes changes in assignment frequency from changes in output weighting.

\paragraph{Differentiation.} The tied configuration $P_1=\dots=P_K$ is symmetric, but it is preserved by a gradient update only when the gradient components of the views coincide. Each $P_m$ receives gradients through the selected experts that assign nonzero mixing weight to view $m$, scaled by $a_{m,i}$ in Eq.~\ref{eq:mv}. Because these weights can vary across experts and tokens, the views can receive different updates without an explicit diversity penalty. We assess router disagreement (\S\ref{sec:view}) and per-view load (App.~\ref{app:view_load}).
The checkpoint probes in \S\ref{sec:view} measure mixing-weight and expert-input diversity. Controlled interventions in App.~\ref{app:mv_probes} test whether the trained model uses the correspondence between experts, tokens, and mixtures.

\section{Experimental Details and Additional Analyses}
\label{app:extra}

\subsection{Full configuration}
\label{app:setup}
Appendix~\ref{app:arch10b} gives the complete architecture and training configuration. The model has hidden dimension $D{=}1024$, 24 layers, latent dimension $d{=}512$, and $N{=}256$ routed experts with top-8 selection and one always-active shared expert. Matrix parameters use Muon, while embeddings, the output head, biases, and router weights use AdamW. Training uses a warmup--stable--decay schedule with peak learning rate $6\times10^{-4}$, 4096-token sequences, approximately $7.9$M tokens per optimizer step, and BF16 with FP8 (E4M3) matrix multiplications under 8-way expert parallelism. Each variant in \S\ref{sec:exp} changes one experimental setting.

\subsection{The 10B-A1B configuration}
\label{app:arch10b}
Table~\ref{tab:arch10b} specifies the 10B-A1B latent-space MoE, with approximately 10B total and 1B activated parameters. This configuration is used for all main results, ablations, and diagnostics in \S\ref{sec:exp}.

\begin{table}[t]
\centering
\small
\setlength{\tabcolsep}{4pt}
\resizebox{\columnwidth}{!}{%
\begin{tabular}{ll}
\toprule
Component & Setting \\
\midrule
Layers / hidden size $D$ & 24 / 1024 \\
Normalization / embeddings & RMSNorm / untied \\
\midrule
Attention & MLA, 16 attention heads, gated output \\
\quad Query / KV latent rank & 512 / 512 \\
\quad QK dims per attention head (no RoPE + RoPE) & 192 + 64 \\
\quad Value dim per attention head & 256 \\
\quad QK-Norm / RoPE base & yes / $10^4$ \\
\midrule
MoE layers & all 24 \\
\quad Latent dim $d$ & 512 \\
\quad Routed experts / active & 256 / 8 \\
\quad Routed expert FFN & SwiGLU, intermediate 1024 \\
\quad Shared experts & 1, intermediate 1024 \\
\quad Balancing & expert bias, rate $10^{-4}$ \\
\midrule
MTP & 1 module, dense FFN 5120, weight 0.3 \\
\bottomrule
\end{tabular}}
\caption{Architecture of the 10B-A1B model.}
\label{tab:arch10b}
\end{table}

\paragraph{Attention.} Each layer uses multi-head latent attention (MLA) \citep{deepseekai2024deepseek} with 16 attention heads. Queries are compressed to rank 512, and keys and values share a rank-512 latent representation. Each attention head has 192 query--key dimensions without rotary embeddings, 64 decoupled rotary dimensions, and a 256-dimensional value. QK-normalization is applied, rotary embeddings use base $10^4$, and the attention output is gated. The cache stores the 512-dimensional KV latent and the 64-dimensional rotary key for each token and layer.

\paragraph{MoE.} All 24 FFN layers are latent-space MoE layers \citep{elango2026latentmoe}, with no preceding dense FFN layers. Each token is projected from $D{=}1024$ to $d{=}512$, and 8 of 256 routed experts are activated. Each routed expert is a latent-space SwiGLU FFN with intermediate dimension 1024; one shared expert with the same intermediate dimension is always active. Expert biases are updated at rate $10^{-4}$ for load balancing \citep{wang2024auxiliary}, and the auxiliary-loss coefficient is 0.

\paragraph{MTP.} Following \citet{deepseekai2024deepseekb}, one multi-token prediction module predicts one additional future token. Its FFN is a dense SwiGLU with intermediate dimension 5120. Its loss is added to the language-modeling loss with weight 0.3.

\paragraph{Training.} The sequence length is 4096 tokens, and the global batch contains 1920 sequences, or approximately 7.9M tokens per step. Matrix parameters use Muon \citep{liu2025muon} with momentum 0.95 and five Newton--Schulz steps. Embeddings, the output head, biases, and router weights use AdamW with $\beta=(0.9,0.95)$ and $\epsilon=10^{-15}$. The learning rate reaches $6\times10^{-4}$ after 1000 warmup steps and follows a warmup--stable--decay schedule ending at 0. Weight decay is 0.1, and gradients are clipped at 1.0. Training uses BF16 with FP8 (E4M3) matrix multiplications, 8-way expert parallelism, and seed 42. The main baseline, Multi-Router, and Multi-View comparisons use the 708B-token checkpoints; training ablations use a common checkpoint.

\subsection{Comparability protocol}
\label{app:repro}
The loss differences are of order $10^{-3}$, making configuration consistency particularly important. We use the following comparison protocol.

\emph{Run-pair reference.} A pair of near-matched runs exhibits a systematic loss offset. Its observed difference provides an empirical reference for the ablation magnitudes in Table~\ref{tab:main}.

\emph{Matched code and configuration.} Each variant uses a common code snapshot with one experimental setting changed. Comparisons use configurations recorded during training, including initialization and its scale, optimizer, schedule, and data order.

\emph{Complementary diagnostics.} We report expert load and utilization (App.~\ref{app:view_load}) alongside loss to characterize how the expert pool is used.

\subsection{Additional measurements}
\label{app:more}

\paragraph{Downstream protocol.} We evaluate the baseline, \mr{}, and score-derived \mv{} checkpoints from training step 90K, corresponding to 708B tokens at 7.86M tokens per step, on the nine tasks in Table~\ref{tab:bench}. CMMLU \citep{li2023cmmlu}, MMLU \citep{hendrycks2020mmlu}, MMLU-Redux \citep{gema2024mmluredux}, MMLU-Pro \citep{wang2024mmlupro}, and BBH \citep{suzgun2022bbh} use 5-shot prompts. GPQA \citep{rein2023gpqa} uses 10-shot prompts. These tasks are answered without chain of thought. HotpotQA \citep{yang2018hotpotqa}, SimpleQA \citep{wei2024simpleqa}, and PreciseWikiQA \citep{bang2025hallulens} use 10-shot prompts with chain of thought and are scored by an LLM judge (DeepSeek-V4-Flash). Group means and the overall mean are unweighted.

\begin{figure}[t]
\centering
\includegraphics[width=\columnwidth]{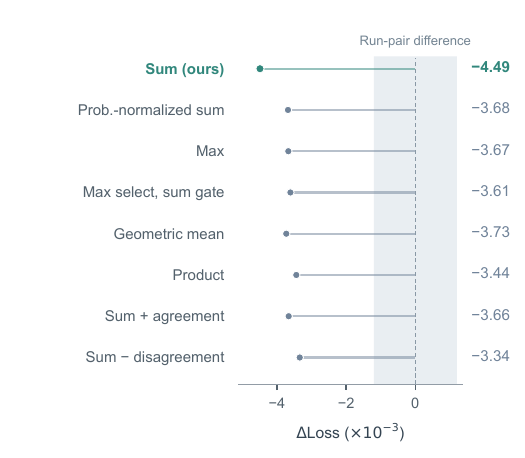}
\caption{Training-loss changes for router-fusion variants with $R=2$. All evaluated variants improve over the single-router baseline. The shaded range shows the observed run-pair difference.}
\label{fig:operators}
\end{figure}

\paragraph{Training trajectories and aggregation operators.}
Figure~\ref{fig:loss} in the main text shows the training trajectories of the baseline, Multi-Router, and Multi-View through 708B tokens and their loss differences. Figure~\ref{fig:operators} compares the post-sigmoid fusion operators at the shared ablation checkpoint.

\subsection{Router-score alignment}
\label{app:router_diagnostics}
For a layer and a sampled token set $T$, let $z_t$ be the latent input of token $t$. Evaluating all $N=256$ experts gives the normalized output-norm proxy
\begin{equation}
\begin{aligned}
q_{t,i}&=\lVert E_i(z_t)\rVert_2,\\
\mu_i&=\frac{1}{|T|}\sum_{t\in T}q_{t,i},\qquad
u_{t,i}=\frac{q_{t,i}}{\mu_i}.
\end{aligned}
\label{eq:router_proxy}
\end{equation}
The normalization removes each expert's average output scale. Let $s_{t,i}$ be the score before balancing bias: the score from a single routing head for the baseline and the summed scores from all routing heads for Multi-Router. Correlation is computed over the $N$ experts for each token:
\begin{equation}
\rho_t=\mathrm{Corr}\!\left(\mathrm{rank}(s_t),\mathrm{rank}(u_t)\right).
\label{eq:router_rho}
\end{equation}
Token values are averaged within each microbatch to obtain $\rho_{\ell,p,b}$ for layer $\ell$, rank $p$, and microbatch $b$. With participating ranks $\mathcal P$ and microbatches $\mathcal B_{\ell,p}$, the layer and global averages are
\begin{equation}
\begin{aligned}
\rho_\ell&=\frac{1}{|\mathcal P|}\sum_{p\in\mathcal P}
\frac{1}{|\mathcal B_{\ell,p}|}\sum_{b\in\mathcal B_{\ell,p}}\rho_{\ell,p,b},\\
\rho_{\mathrm{global}}&=\frac{1}{24}\sum_{\ell=1}^{24}\rho_\ell.
\end{aligned}
\label{eq:router_aggregation}
\end{equation}
For layer-wise summaries, we match the two diagnostic exports by logged step and average the last 201 records available in both exports. Layer-wise bands show one standard deviation over these measurements.

\subsection{Expert-input diversity and controlled mixing interventions}
\label{app:mv_probes}

\paragraph{Evaluation protocol.}
All probes in this subsection use the score-derived two-view checkpoint at step 90K, the same checkpoint used in Table~\ref{tab:bench}. The main evaluation contains 256 held-out sequences of 4096 input tokens, giving 1,048,320 next-token predictions; the clean checkpoint's NLL is $2.17617$ nats/token. The layer scan uses 80 sequences, giving 327,600 predictions. No model parameters are updated.

\begin{figure*}[t]
\centering
\includegraphics[width=\textwidth]{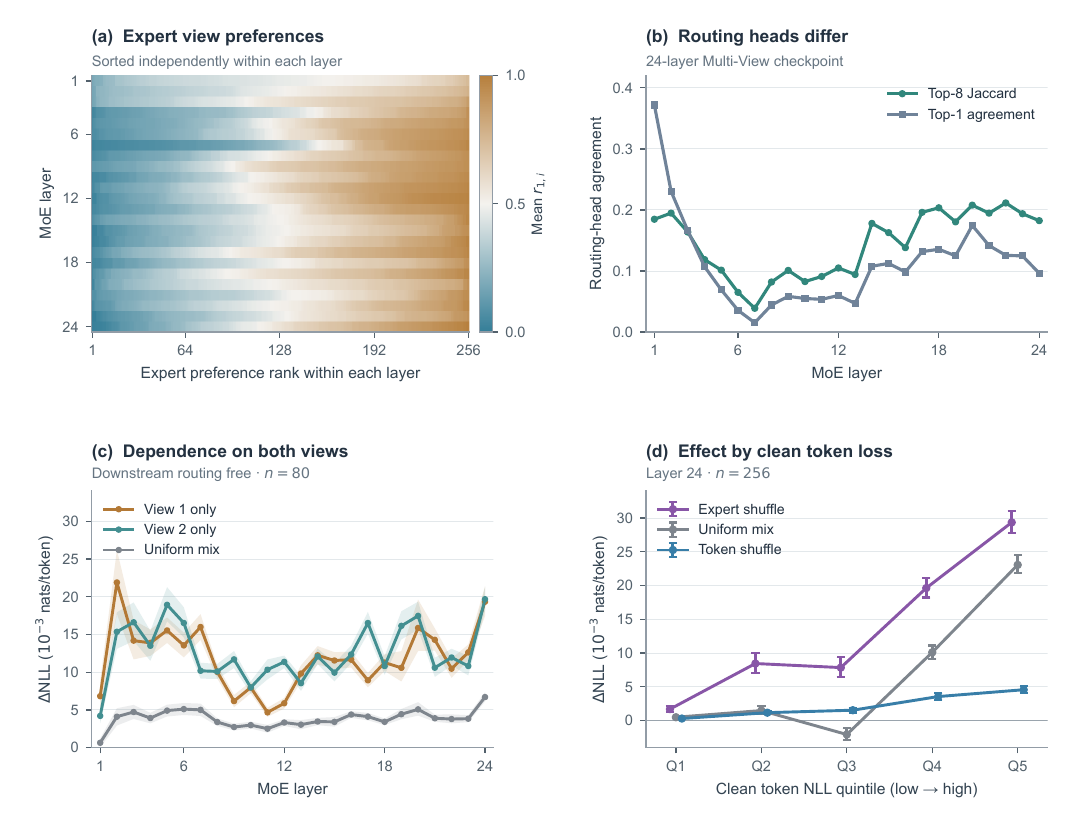}
\caption{Additional analyses of the 708B-token \mv{} checkpoint. \textbf{(a)} Mean view-1 share $r_{1,i}$ by expert, with expert ranks determined independently in each layer. \textbf{(b)} Agreement between the two routing heads on 256 sequences. \textbf{(c)} Single-view interventions on 80 sequences, with downstream routing recomputed; bands are pointwise 95\% CIs. \textbf{(d)} Final-layer effects grouped by clean token-NLL quintile on 256 sequences. The x-axis indexes five groups with equal numbers of predictions; error bars show 95\% intervals.}
\label{fig:mv_supplement}
\end{figure*}

\paragraph{Diversity definitions.}
For token $t$, let $\mathcal{S}_t$ be its $k=8$ selected experts. The probe uses the main method's actual mixing proportions, $r_{m,i}(t)=a_{m,i}(t)=s_{m,i}(t)/\sum_n s_{n,i}(t)$, so $\sum_m r_{m,i}(t)=1$ and $r_{2,i}(t)=1-r_{1,i}(t)$. The actual input to expert $i$ is
\[
\hat v_i(t)=r_{1,i}(t)v^{(1)}(t)+r_{2,i}(t)v^{(2)}(t).
\]
We compute
\begin{align}
D_r(t)&=\frac{2}{k(k-1)}\sum_{\substack{i<j\\i,j\in\mathcal{S}_t}}|r_{1,i}(t)-r_{1,j}(t)|,\\
D_v(t)&=\frac{\frac{2}{k(k-1)}\sum_{\substack{i<j\\i,j\in\mathcal{S}_t}}\|\hat v_i(t)-\hat v_j(t)\|_2}{\frac{1}{k}\sum_{i\in\mathcal{S}_t}\|\hat v_i(t)\|_2}.
\end{align}
The bands in Fig.~\ref{fig:mv_mechanism}(a) are the 25th--75th percentiles across tokens. Mean cosine similarity between the two views ranges from $0.059$ to $0.530$ across layers. The median view-separation norm, divided by the mean of the two view norms, ranges from $0.969$ to $1.380$. The projections therefore produce distinct views at this checkpoint.

\paragraph{Expert preferences.}
Fig.~\ref{fig:mv_supplement}(a) shows each expert's mean $r_{1,i}$ over tokens selecting it; experts are sorted independently within each layer. All 256 experts are observed in each layer. The within-token component accounts for $84.8$--$88.7\%$ of the measured share variance, while grouping by expert ID explains $28.0$--$73.4\%$. Across layers, top-8 Jaccard between routing heads ranges from $0.039$ to $0.211$, and top-1 agreement ranges from $1.5\%$ to $37.2\%$ (Fig.~\ref{fig:mv_supplement}(b)).

\paragraph{Interventions and uncertainty.}
Expert permutation shuffles the view proportions among the eight selected experts of each token. Token permutation shuffles them among tokens selecting the same expert; $99.96\%$ of selected assignments are permutable. For each selected assignment, a replacement $r_{1,i}^{\prime}$ is implemented through the dispatcher channels $(g_i r_{1,i}^{\prime},g_i(1-r_{1,i}^{\prime}))$. The output gate $g_i$ remains attached to the original assignment and is not permuted. Uniform mixing uses $(g_i/2,g_i/2)$; the two single-view interventions use $(g_i,0)$ or $(0,g_i)$. View projections, expert parameters, the selected expert set, and output gates at the intervened layer are retained. All reported changes use paired predictions on the same sequences. For the final-layer permutation tests, we average three permutation seeds within each sequence and bootstrap sequences to obtain 95\% CIs. Table~\ref{tab:mv_intervention} reports the final-layer results. The probe reports mean absolute changes in the view-1 share of $0.408$, $0.117$, and $0.331$ for expert permutation, token permutation, and uniform mixing, respectively.

\begin{table}[t]
\centering\small
\setlength{\tabcolsep}{3pt}
\begin{tabular}{lrr}
\toprule
Intervention & $\Delta$NLL & 95\% CI\\
\midrule
Identity replay & $0.0003$ & $[-0.001,0.002]$\\
Token permutation & $2.23$ & $[2.05,2.41]$\\
Uniform mixture & $6.62$ & $[6.20,7.09]$\\
Expert permutation & $13.40$ & $[12.56,14.35]$\\
View 1 only & $18.81$ & $[17.58,20.12]$\\
View 2 only & $20.42$ & $[19.32,21.56]$\\
\bottomrule
\end{tabular}
\caption{Final-layer interventions on 256 held-out sequences. All values, including CIs, are in $10^{-3}$ nats/token. Expert selection and output gates $g_i$ are fixed to numerical precision; only view-mixing proportions change.}
\label{tab:mv_intervention}
\end{table}

\paragraph{Execution and replay checks.}
On a 16-sequence execution-check subset, the dispatcher receives the same expert IDs across interventions; output gates differ by at most $1.2\times10^{-7}$ in relative error, and the realized view proportions at the dispatcher differ from their intended values by at most $6.0\times10^{-8}$. Two clean passes agree exactly on 32 sequences. At layer~24, identity replay gives a mean NLL change of $2.95\times10^{-7}$ nats/token, with a 95\% CI spanning zero, and a mean absolute token-level difference of $1.14\times10^{-5}$. Using replay rather than the clean pass as the paired reference leaves the reported effects unchanged at the displayed precision.

For shallower interventions, rounding-level changes can alter later top-$k$ decisions. At layer~1, identity replay changes the final-layer expert set for $6.3\%$ of tokens in the execution check. On the 80-sequence scan, its mean NLL change is $5.51\times10^{-6}$, with 95\% CI $[-1.37,1.57]\times10^{-4}$ nats/token. The scan estimates total effects with downstream routing recomputed. Its intervals use its own 80-sequence sample and are pointwise intervals for individual layers.

\paragraph{Single-view and token-group results.}
Both single-view interventions increase NLL at every scanned layer (Fig.~\ref{fig:mv_supplement}(c)); their final-layer effects on 256 sequences appear in Table~\ref{tab:mv_intervention}. Both branches contribute to prediction in this trained checkpoint. Fig.~\ref{fig:mv_supplement}(d) groups predictions into equal-sized quintiles of clean token NLL, with 209,664 predictions per group. Expert permutation has its largest absolute effect in the highest-loss quintile ($0.02936$ versus $0.00170$ in the lowest). Uniform mixing gives a middle-quintile $\Delta$NLL of $-0.00206$. The largest intervention effects occur in the highest-loss group.

\subsection{Aggregate and per-view expert load}
\label{app:view_load}
Per-view load measurements show how each view's recorded weight mass is distributed across the shared expert pool. Load balancing acts once on the aggregated assignment over the $N$ experts (App.~\ref{app:impl}), so different views may concentrate on different subsets while the aggregate load remains balanced. We therefore report, for each view, the coefficient of variation (CV) of its recorded weight mass across experts and an effective expert count $\exp(H_m)$, where $H_m$ is the entropy of this normalized weight mass (App.~\ref{app:props}). The aggregate load CV is $0.854$, compared with $0.862$ for the baseline, and the aggregate effective count is $188.7$, compared with $187.6$ out of $256$. The individual views have CVs of $1.155$ and $1.144$ and effective counts of $149.0$ and $150.2$. These counts correspond to approximately $58\%$ of the pool for each view and $74\%$ in aggregate. This pattern occurs across MoE layers (layer 2: $218$ in aggregate versus $178$ and $165$; layer 15: $159$ versus $123$ and $130$). Together, the views combine concentrated expert preferences with broader aggregate expert usage.

\end{document}